\documentclass[10pt,twocolumn,letterpaper]{article}
\usepackage[dvipsnames]{xcolor}

\usepackage[pagenumbers]{cvpr} 
\usepackage[ruled,vlined]{algorithm2e}
\usepackage{graphicx}
\usepackage{amssymb} 
\usepackage{amsmath}
\usepackage{color}
\usepackage{tikz}
\usepackage{comment}
\usepackage{graphicx}
\usepackage{amsmath}
\usepackage{amssymb}
\usepackage{float}
\usepackage{multirow}
\usepackage{soul}
\usepackage{subcaption}
\usepackage{pifont}
\newcommand{\cmark}{\ding{51}}%
\newcommand{\xmark}{\ding{55}}%

\definecolor{cvprblue}{rgb}{0.21,0.49,0.74}
\usepackage[pagebackref,breaklinks,colorlinks,citecolor=cvprblue]{hyperref}

\def\paperID{*****} 
\def\confName{CVPR}
\def\confYear{2024}

\title{Dual Guidance Semi-Supervised Action Detection} 
\author{Ankit Singh$^{1}$\ \ \ \ Efstratios Gavves$^{2}$  \ \ \ \ Cees G. M. Snoek \ \ \ \ Hilde Kuehne$^{3,4}$ \\ \\
$^{1}$ IIT Madras, $^{2}$ University of Amsterdam,$^{3}$ University of Tuebingen, $^{4}$ MIT-IBM Watson AI Lab
}

\begin{document}
\maketitle

\begin{abstract}
Semi-Supervised Learning (SSL) has shown tremendous potential to improve the predictive performance of deep learning models when annotations are hard to obtain. However, the application of SSL has so far been mainly studied in the context of image classification. In this work, we present a semi-supervised approach for spatial-temporal action localization. We introduce a dual guidance network to select better pseudo-bounding boxes. It combines a frame-level classification with a bounding-box prediction to enforce action class consistency across frames and boxes. 
Our evaluation across well-known spatial-temporal action localization datasets, namely UCF101-24 , J-HMDB-21 and AVA shows that the proposed module considerably enhances the model's performance in limited labeled data settings. Our framework achieves superior results compared to extended image-based semi-supervised baselines.

\end{abstract}
\section{Introduction}
Spatial-temporal localization ~\cite{Kalogeiton2017ActionTD,Yang2019STEPSP,Kpkl2019YOWO,Li2020ActionsAM} of actions in video data is a challenging task, as it requires combining spatial detection, temporal segmentation, as well as action classification. Defining the spatial and temporal extent of action can be challenging even for human annotators. In action detection, it is not enough to learn the pure outline of a figure, but also to capture motion information, ~\cite{Zolfaghari2017ChainedMN,Tomei2021VideoAD,Gavrilyuk2021MotionAugmentedSF}, environmental cues ~\cite{Chiou2021STHOIAS} and more to come to the correct prediction. 
Deep learning together with the availability of large-scale data has led to a significant performance improvements across various vision tasks such as action recognition \cite{Wang2019TemporalSN,Lin2019Tsm,Feichtenhofer2019Slowfast,Simonyan2014TwoStreamCN} , object detection \cite{Ren2015FasterRT,redmon2017yolo9000,lin2017focal,Lin2017FeaturePN} and action detection ~\cite{Mettes2021ObjectPF,Zhao2017TemporalAD,Zhao2021TubeRTT,Mettes2017LocalizingAF}. However, the success of those frameworks usually depends on the availability of a large amount of labeled data and did so far not translate to applications such as spatial-temporal localization, where annotating data is highly time-consuming as it involves the annotator going through all the frames of the video to label the corresponding bounding boxes. 

Semi-supervised learning (SSL) tries to overcome the problem of annotation scarcity by relying only on a small set of labeled instances which can be used to leverage a larger set of available unlabeled data. The semi-supervised approaches have shown tremendous potential in the image domain, even surpassing the performance achieved by their supervised counterparts for various tasks such as image recognition ~\cite{Tarvainen2017Mean,berthelot2019remixmatch,Berthelot2019Mixmatch,Sohn2020Fixmatch} and object detection ~\cite{Tang2021HumbleTT,Yang2021InteractiveSW,Liu2021UnbiasedTF,Xu2021EndtoEndSO}. 
This motivates us to leverage the idea of SSL to address the problem of spatial-temporal action localization.

Moreover, in order to explore the impact of limited annotations on supervised network training, we analyzed the localization and classification recall of the network. Localization recall is defined as the number of predicted frames having $IoU > 0.5$ over total ground truth frames, whereas classification recall is denoted as the number of predicted frames having $IoU > 0.5$ with the correct class label over total ground truth frames. We found a huge gap between these two measures in limited annotation settings. For instance, a network trained on UCF101-24 with pretrained weights has $76.1\%$ localization recall compared to $59.16\%$ classification recall, and the gap further widens to have $26.5\%$ localization recall and $9.6\%$ classification recall when the network is trained without any pretrained weights. 
To this end, we propose a dual guidance framework that combines the local and global characteristics of spatial-temporal localization in a semi- supervised setup. This helps us to better select the pseudo-bounding boxes for training unlabeled videos.

Overall, we follow the lines of \cite{Sohn2020Fixmatch,Tarvainen2017Mean} and use a model pretrained on the labeled data to make predictions for weak and strongly augmented versions on the unlabeled data to use as pseudo labels to improve the system.
To generate those pseudo-labels, on the local level, we follow the idea of action \cite{Kpkl2019YOWO} and object detection \cite{Liu2021UnbiasedTF,Wang2021DataUncertaintyGM,Zhou2021InstantTeachingAE} and try to predict a bounding box for each detected action in each frame. 
On the global level, we leverage an action classifier based on the whole frame information to predict the overall class label.
The global action labels are then used to support or reject local bounding box proposals and refine the pseudo-bounding box selection. 
This can be helpful as the global system can capture additional information on background cues such as indoor or outdoor settings, nearby objects, and others that might be relevant for classification, while the local system can focus on the local appearance. 
Finally, we further include a temporal supervision with respect to global frame categories. 
We implement this as, while the location of a bounding box can significantly change within a few frames, the action category of the video clip should remain consistent across frames.

We evaluate our framework on the well spatial-temporal action localization dataset, UCF101-24, J-HMDB-21 and AVA dataset which demonstrate that our framework achieves superior performance over extended baselines of the state-of-the-art semi-supervised images domain approaches.
In summary, our key contributions are as follows:
\begin{itemize}
    \item We propose a novel, simple single-stage training
    framework for action detection in real-world video data.
    \item We propose a dual guidance-based method for semi-supervised action detection combining local bounding box and global frame-level supervision to select the pseudo-bounding boxes.
    \item We provide an extensive evaluation of this new problem and demonstrate the framework's effectiveness over the extended image-domain semi-supervised approaches. We also show the in-depth analysis of our framework, highlighting the impact of each component. 
\end{itemize}


\section{Related Works}
\subsection{Spatial-Temporal Action Localization }
Recently, spatial-temporal action localization has drawn attention from the community, and new datasets are being introduced. Conventional approaches used by early works ~\cite{Girdhar2019VideoAT,Yang2019STEPSP,Li2018RecurrentTP} on the problem used R-CNN detectors on 3D-CNN features. Frame-level~\cite{Gkioxari2015FindingAT,Peng2016MultiregionTR,Saha2016DeepLF,Singh2017OnlineRM,Weinzaepfel2015LearningTT} and clip-level detection ~\cite{Gan2015DevNetAD,Wang2016ActionnessEU,Hou2017TubeCN,Yang2019STEPSP,Li2018RecurrentTP,Zhao2019DanceWF,Song2019TACNetTC} are most  explored approaches in the spatial-temporal action localization. ~\cite{Gkioxari2015FindingAT,Peng2016MultiregionTR,Saha2016DeepLF,Singh2017OnlineRM,Weinzaepfel2015LearningTT} use frame-level detection and linking algorithm ~\cite{Gkioxari2015FindingAT,Hou2017TubeCN} to generate final action tubes. ACT ~\cite{Kalogeiton2017ActionTD} used clip level detection to make action tublets regress over the anchor cuboids. STEP ~\cite{Yang2019STEPSP} progressively refine the proposals to utilize the longer temporal information present in videos. MOC ~\cite{Li2020ActionsAM} proposed anchor-free tublet detection where action sequences are treated as a trajectory of moving points. Recently graph-structured networks ~\cite{Tomei2019STAGESA,Sun2018ActorCentricRN,Zhang2019ASM} have been proposed to exploit the contextual information. TubeR~\cite{Zhao2021TubeRTT} directly predicts an action tublet in a video performing action localization and recognition simultaneously. However, all these approaches are supervised and require a fully annotated dataset. Due to huge annotation requirements in such a task, the community has turned towards a weakly supervised setting ~\cite{Escorcia2020GuessWA,Li2018VideoLSTMCA,Mettes2017LocalizingAF} where only a video-action label is available. In contrast to the above, our work proposes a semi-supervised framework for Spatial-Temporal Action Localization.

\subsection{Semi-Supervised Learning}
Semi-Supervised learning aims to leverage the vastly available unlabeled data to improve the model's performance. Recently many semi-supervised learning approaches 
\cite{Tarvainen2017Mean,Sohn2020Fixmatch,Berthelot2019Mixmatch,Zhai2019S4L,berthelot2019remixmatch} have shown tremendous potential and adapted across various other tasks such as domain adaptation and semantic segmentation. Most of these SSL methods try to ensure consistent representation across different transformed views of the images. Mean teacher ~\cite{Tarvainen2017Mean} approach makes two copies of the model one is called teacher while the other is called student model. The approach ensures consistent prediction across both these models while the teacher model is the student model's EMA. Mix-Match ~\cite{Berthelot2019Mixmatch} and Remix-Match ~\cite{berthelot2019remixmatch} use interpolation between labeled and unlabeled data as a practical data augmentation approach to improving the model's performance.
Recently introduced Fix-Match~\cite{Sohn2020Fixmatch}, which uses a simple method of considering the highly confident pseudo labels as the ground truth for the unlabeled data, this simple yet effective strategy can be used across a variety of tasks.
However, these approaches have been developed for image-classification tasks where we deal with spatial-temporal action localization task in our work.

\subsection{Semi-Supervised Object Detection}
Object detection is one of the essential computer vision tasks and has garnered enormous attention. While a significant problem, it is also where annotating data is time-consuming and expensive. Recent works ~\cite{Liu2021UnbiasedTF,Zhou2021InstantTeachingAE,Wang2021DataUncertaintyGM,Tang2021HumbleTT} introduced strategies to use unlabeled data and limited labeled data to improve the performance of object detectors. Most of these approaches either use pseudo-label approaches ~\cite{Tang2021HumbleTT,Liu2021UnbiasedTF,Zhou2021InstantTeachingAE} or consistency-based approaches ~\cite{Tang2021ProposalLF,Jeong2019ConsistencybasedSL} to train the object detectors with unlabeled data.
\cite{Jeong2019ConsistencybasedSL} ensures the consistency between unlabeled labeled image and its flipped counterpart for semi-supervised object detection. \cite{Radosavovic2018DataDT} uses data distillation to generate labels by aggregating predictions from different transformed views of the unlabeled data.\cite{Sohn2020ASS} uses a pretrained detector to generate pseudo labels on the unlabeled samples, which remain fixed throughout the training and are used to finetune the network. In contrast to these works, which deal with semi-supervised object detection settings, our work aims to propose a better semi-supervised method for Spatial-Temporal Action Localization.

\section{Methodology}
This section will present our novel Semi-Supervised Learning (SSL) based framework for Semi-Supervised Spatial-Temporal Action Localization. We will first discuss the problem and notation used in our work and then describe our framework and its components in detail.
\subsection{Problem Description}
Our work aims to address the spatial-temporal action localization using a small amount of labeled Videos $\mathcal{D}_l$ and large number of unlabeled videos $\mathcal{D}_u$ without any corresponding annotations. The labeled set $\mathcal{D}_l \triangleq \{V^i,Y^i\}_{i=1}^{N_l}$ contains of $N_l$ videos where $V^i$ represents the $i^{th}$ video and $Y^i=\{gt^i, y^i\}$ represents the locations($gt^i$), and action categories $y^i$ of all bounding boxes across different frames of the $i^{th}$ video. Similarly, the unlabeled set $\mathcal{D}_u \triangleq \{U^i\}_{i=1}^{N_u}$ comprises of $N_u (\gg\!\! N_l)$ videos without any associated bounding boxes.

\subsection{Supervised Training}
\begin{figure}
    \centering
    \includegraphics[width=\columnwidth]{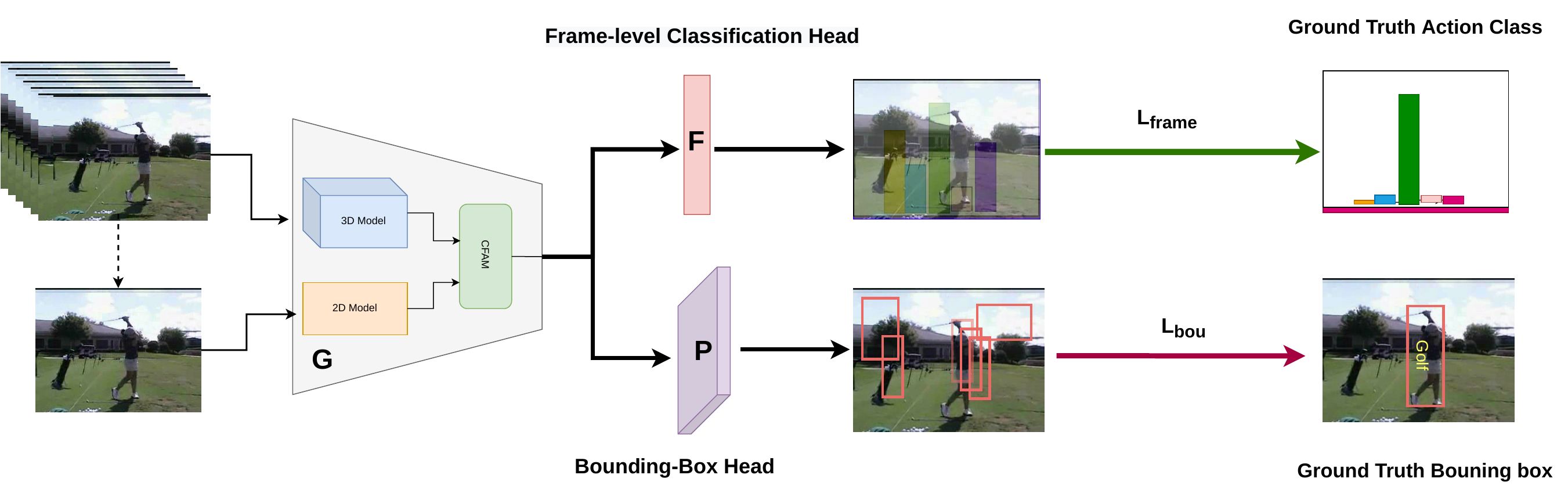}
    \caption{\textbf{Overview of the Supervised Training}: The input clip is passed through $\mathcal{G}$ to extract features, which are then processed through $\mathcal{F}$ and $\mathcal{P}$ where $\mathcal{F}$ is the frame-level classification head and $\mathcal{P}$ is the bounding box head. The prediction from $\mathcal{F}$ is trained against the ground-truth action class while the bounding boxes predicted through $\mathcal{P}$ are trained against the labeled ground truth bounding box.}
    \label{fig:Supervised}
\end{figure}

Our framework involves two stages of training. At the first stage, we train the network with labeled videos using supervised learning before introducing the unlabeled videos at the second stage. During supervised training, labeled videos are passed through the CNN-based feature extractor $\mathcal{G}(.)$ to obtain
corresponding features, which are then passed through Bounding-Box head $\mathcal{P}(.)$  and Classifier $\mathcal{F}(.)$ to obtain corresponding features as shown in fig ~\ref{fig:Supervised}. The bounding-box head $\mathcal{P}(.)$ uses $1 \times 1$ kernels to predict the locations, sizes, and action categories of the bounding boxes. Bounding-Box head predicts action class conditional scores along with four coordinates and confidence score for each grid cell in the output volume similar to YOLO ~\cite{redmon2017yolo9000} architecture. Following ~\cite{girshick2015fast,Kpkl2019YOWO,redmon2017yolo9000}, we used $L1$ loss (\ref{eq:l1loss}) for the localization and $L2$ loss (\ref{eq:l2loss}) for the confidence score of the bounding boxes.

\begin{equation}
    \mathcal{L}_{1} =  
    \begin{cases} 
    \eta(x^i-gt^i)^2  &   if |x^i-gt^i|<1 \\
    |x^i-gt^i|-\eta   &   otherwise
    \end{cases}
    \label{eq:l1loss}
\end{equation}

where $x$ refers to the network prediction ($\mathcal{P(G(}V^i))$) and $gt$ denotes the ground truth available with the labeled data. Following \cite{Kpkl2019YOWO}, we also used
focal loss \cite{lin2017focal} for the bounding box classification as shown in (\ref{eq:focalloss})

\begin{equation}
    \mathcal{L}_{conf}(x,y) = {(x-y)}^2
    \label{eq:l2loss}
\end{equation}

\begin{equation}
    \mathcal{L}_{focal}(x,y) = y(1-x)^\gamma log(x) + (1-y)x^\gamma log(1-x)
    \label{eq:focalloss}
\end{equation}
where $x$ is the predicted probability from the network output while $y$ is the ground truth action class of the bounding box. $\gamma$ is the modulating factor which increases the loss for hard samples and decreases the loss for weak samples. The final bounding box loss $\mathcal{L}_{bou}$ becomes the summation of individual coordinate losses for bounding-box locations, width and height; and confidence score loss and bounding box classification loss which is formulated as follows:
\begin{equation}
    \mathcal{L}_{bou} = \alpha*(\mathcal{L}_x + \mathcal{L}_y + \mathcal{L}_w + \mathcal{L}_h + \mathcal{L}_{conf})+ \mathcal{L}_{focal}
\end{equation}
where $\alpha$ is the weight factor for the coordinate losses and confidence score loss.
Features extracted from $\mathcal{G}(.)$ are also passed through the frame-level classification head $\mathcal{F}(.)$ to predict the action being performed in the frame by minimizing the well-known cross-entropy loss.
\begin{equation}
   \mathcal{L}_{frame} = -\sum\limits_{c=1}^{C+1}(y^i)_c \log(g(V^i))_c
   \label{eq:cross_entropy}
\end{equation}
where $C+1$ denotes the total action categories in the dataset including background. In case of multi-action AVA dataset, we used weighted binary cross-entropy loss.
The total supervised loss is the summation of the bounding box losses $\mathcal{L}_{bou}$ and frame classification loss $\mathcal{L}_{frame}$.
\begin{equation}
    \mathcal{L}_{sup} = \mathcal{L}_{bou} + \beta*\mathcal{L}_{frame} 
\end{equation}

\begin{figure*}[t]
    \centering
    \includegraphics[width=\textwidth]{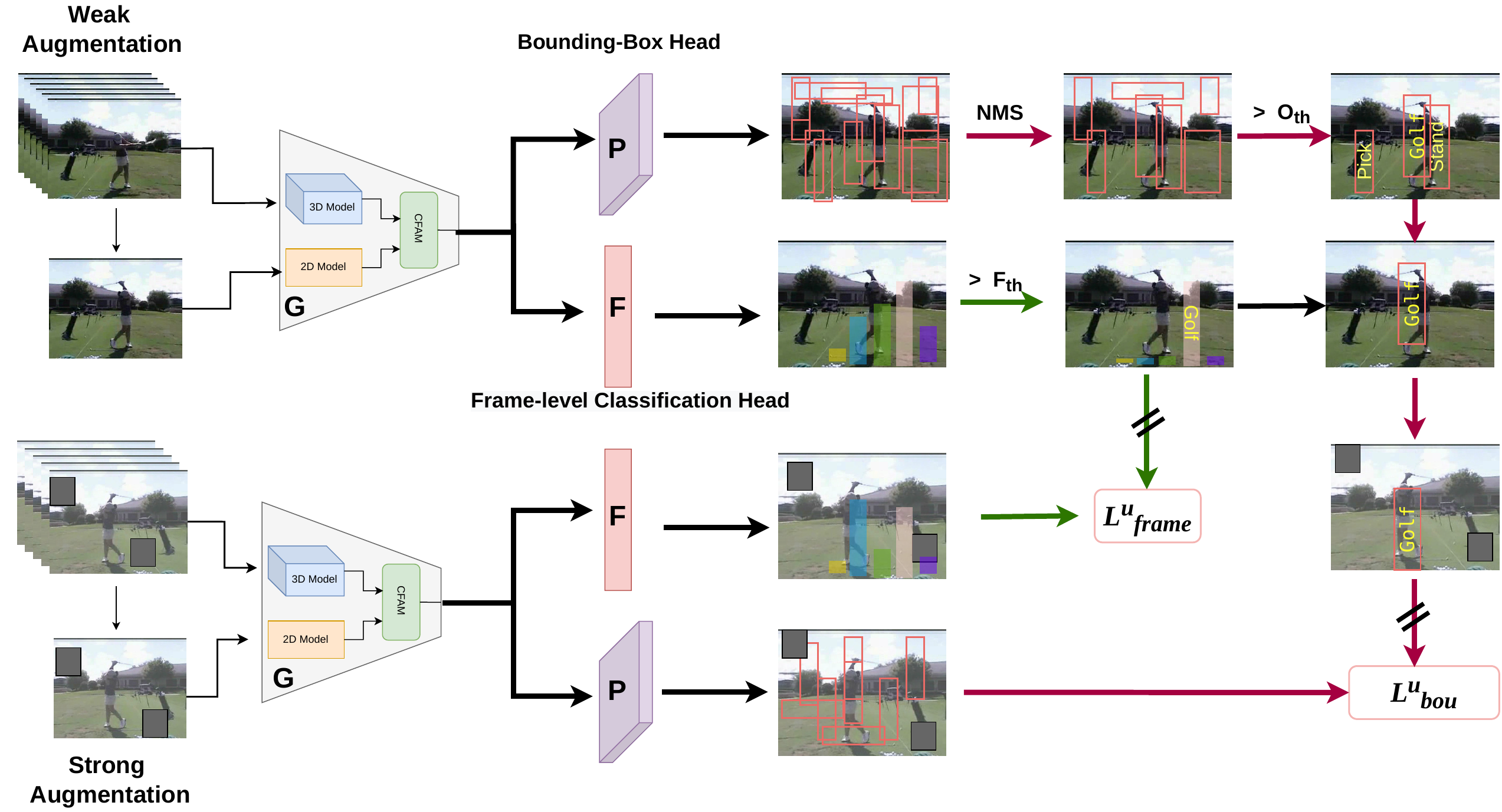}
    \caption{\textbf{Outline of our Framework}: Our framework consists of aligning the outputs of the network at two levels. On the global level, if the frame-level prediction is greater than $F_{th}$, then one hot pseudo-label is generated, and prediction on strongly augmented input is ensured to match the one-hot pseudo-label using cross-entropy loss. On the local level, the features from $\mathcal{G}(.)$ are passed to $\mathcal{P}(.)$ generate the bounding boxes with a class-confidence score. The generated bounding boxes are passed through NMS, and then bounding boxes with confidence scores greater than $O_{th}$ are selected as candidate bounding boxes. The candidate bounding boxes having the same action class as the global action category from $\mathcal{F}(.)$ are considered pseudo-bounding boxes. The bounding boxes generated through $\mathcal{P}$ on the strongly augmented clip are trained against the pseudo bounding boxes using $\mathcal{L}_{bou}$ after localization changes are done as per strong augmentation.}
    \label{fig:overview}
\end{figure*}

\subsection{ Semi-Supervised Learning}
Once the network is trained with labeled data for some epochs, we move into the second stage, introducing the unlabeled data. Since the unlabeled videos do not have any corresponding annotations with them, we use the idea from ~\cite{Sohn2020Fixmatch,Tarvainen2017Mean} to generate the pseudo-labels and ensure the consistent prediction across different augmentation of the input.

To this end, we first create two copies of the model \cite{Tarvainen2017Mean} called the teacher and student model. The teacher and student model are fed with two different augmentations of the input video. Input to the teacher model has weak augmentation( $U^i$) while the student model is fed with strongly augmented input ($\hat{U}^i)$.
The output feature $G(U^i)$ from the teacher model is passed to the frame-level classification head $\mathcal{F}(.)$ and Bounding-Box head $P(.)$. The global action category predicted from $\mathcal{F}(.)$ is considered as pseudo-action label if the maximum probability of the predicted class \textit{i.e} $max(\mathcal{F(G(}U^i))$ is greater than threshold $F_{th}$. This pseudo action label( $\hat{y}_u^i =  argmax(\mathcal{F(G(}U^i)))$) is considered as the ground truth for $\mathcal{F(G(}\hat{U}^i))$ \textit{i.e} output generated from $\mathcal{F(.)}$ for the strongly augmented input. The loss to train the frame-level classification head is defined in (\ref{eq:fixmatch}).

\begin{equation}
\mathcal{L}_{frame}^{u}\!\! = \!\!\!-\sum\limits_{b=1}^{B}\!\!\!\!\mathbf{1} (max(\mathcal{F(G(}U^i))\!\!>\!\! F_{th})H(\mathcal{F(G(}\hat{U}^i)),\hat{y}_u^i)  
   \label{eq:fixmatch}
\end{equation}
where $B$ is the batch size and $H$ is the cross-entropy formulation as described in (\ref{eq:cross_entropy}).
The features generated by $\mathcal{G}(U^i)$ for weakly augmented input clip are also passed through $\mathcal{P(G}(U^i))$ to generate the possible bounding boxes, which are then passed through Non-Maximum Suppression (NMS). The bounding boxes having a confidence score greater than $O_{th}$ becomes the candidate bounding boxes. The candidate bounding boxes are matched with the pseudo-action class prediction from the $F(.)$. The bounding boxes that pass the above criteria are considered pseudo-bounding boxes. These pseudo-bounding boxes act as the ground truth bounding boxes for the predictions of the strongly augmented input clip. The unsupervised version of $L_{bou}$ is used here for training the network with unlabeled videos clip, which can be formulated as follows:

\begin{equation}
    \mathcal{L}_{bou}^{u} = \\
     \alpha*(\mathcal{L}_x^{u} + \mathcal{L}_y^{u} + \mathcal{L}_w^{u} + \mathcal{L}_h^{u} + \mathcal{L}_{conf}^{u})+ \mathcal{L}_{focal}^{u}
\end{equation}
where individual losses are unsupervised version of their counterparts and ground truth available in supervised version is replaced with pseudo-ground truth obtained from pseudo-bounding box selected through the approach described above. $L_{bou}^{u}$ is only computed for $U^i$ unlabeled video clip when  $(\mathcal{F(G(}U^i))>F_{th})$ and $(\mathcal{P(G(}U^i))>O_{th})\quad \land \quad \hat{y}_u^i = argmax(\mathcal{P(G(}U^i))_c )$
where $\mathcal{P(G(}U^i))_c$ represents the class conditional scores of the action categories. The total unsupervised loss is the summation of the $\mathcal{L}_{bou}^{u}$ and $\mathcal{L}_{frame}^{u}$.
\begin{equation}
    \mathcal{L}_{u} = \mathcal{L}_{bou}^{u} + \delta*\mathcal{L}_{frame}^{u} 
\end{equation}
\subsection{Temporal Learning}
While the action localization may vary fast across the frames of the video clip but the global frame-level action representation of the video clip should be consistent as the overall global action category of the continuous frames remain the same.
We used the temporal loss defined in (\ref{eq:ltemporal}) to exploit the temporal consistency present in video clips. We use temporal loss during both phase of training .
\begin{equation}
    \mathcal{L}_{tmp} =  |\mathcal{F(G}(V_{s}^i))-\mathcal{F(G(}V_{s+1}^i))|
    \label{eq:ltemporal}
\end{equation}
where $V_{s}^i$ and $V_{s+1}^i$ represents the $s_{th}$ and $s+1_{th}$ frames of the $i^{th}$ video. 
\subsection{Overall Framework and Training Objectives}
The overall training objectives employs supervised loss, unsupervised losses and temporal losses for the labeled and unlabeled videos which can be formulated as
\begin{equation}
    \mathcal{L}_{total} = \mathcal{L}_{sup} + \mathcal{L}_{u} +\mathcal{L}_{tmp}^{sup} + \mathcal{L}_{tmp}^{unsup}
    \label{eq:total}
\end{equation}

\begin{algorithm}
\DontPrintSemicolon
\SetKwInput{KwInput}{Input}
\SetAlgoLined
\KwInput{Labeled dataset $\{{\mathcal{D}_{l}}\}$, UnLabeled dataset $\{\mathcal{D}_{u}\}$,  and  Model $\{\mathcal{G}, \mathcal{F}, \mathcal{P}\}$}

\For{steps $1$ to $total steps$} {
    Load a mini-batch of labeled samples $\{({V}_{i}, y_{i})\}_{i=1}^{i=B}$ from labeled dataset ${\mathcal{D}_{l}}$ and unlabeled samples $\{({U}_{i})\}_{i=1}^{i=B}$ from Unlabeld dataset ${\mathcal{D}_{u}}$\\
    
    Compute $\mathcal{L}_{bou}$ on labeled samples through $\mathcal{P(G(V}^i))$.\\
    Compute $\mathcal{L}_{frame}$ on labeled samples through $\mathcal{F(G(V}^i))$.\\
    Compute pairwise temporal loss $\mathcal{L}_{tmp}$ on labeled samples\\
    
    Load a mini-batch of unlabeled target samples $\{(\mathbf{U}_{i}\}_{i=1}^{i=B}$ from unlabeled dataset\\
    Create two views of the unlabelled samples, one with augmentation and another without any augmentation.\\
    Compute bounding boxes on unaugmented unlabeled samples through $\mathcal{P(G(V}^i))$.\\
    Pass the generated bounding boxes through NMS\\
    Filter bounding boxes having confidence score $>=$ $O_{th}$\\
    
    Generate Frame level prediction on unaugmented unlabeled samples through $\mathcal{F(G(V}^i))$.\\
    Select samples from mini-batch,$B$ whose class-confidence score $>=$ $F_{th}$ as pseduo frame labels\\
    
    Ensure bounding boxes class prediction matches Frame-level Prediction and selects those bounding boxes as pseudo-bounding boxes\\ 
    
    Pass the augmented samples through the model to generate bounding boxes through $\mathcal{P(G(V}^i))$ and frame-level prediction through $\mathcal{F(G(V}^i))$
    
    Compute $\mathcal{L}_{bou}$ using the generated bounding boxes and pseudo-bounding\\
    Compute $\mathcal{L}_{frame}$ using pseudo-labels generated from the unaugmented samples\\
    
    Compute pairwise temporal loss for the unlabeled samples
    
    Propagate all the losses

    }
\caption{DGA- \textbf{D}ual \textbf{G}uidance \textbf{A}pproach for Semi-Supervised Spatial-Temporal Action Localization}
\label{algo}
\end{algorithm}

\section{Experiments}
In this section, we conduct extensive experiments across different datasets to study the effectiveness of our framework. We also conducted an ablation study to understand the effect of different modules of our framework.
\begin{table}[ht]
\centering
\resizebox{\columnwidth}{!}{%
    \begin{tabular}{c|c|c|c|c}
    \hline
    \multicolumn{1}{c|}{Dataset} & \multicolumn{1}{c|}{\begin{tabular}[c]{@{}c@{}}\# of annotated\\ action per frame\end{tabular}} &
    \multicolumn{1}{c|}{\begin{tabular}[c]{@{}c@{}}\# of action\\ per frame\end{tabular}} &
    \multicolumn{1}{c|}{\begin{tabular}[c]{@{}c@{}}background\\ people \end{tabular}}   &
    \multicolumn{1}{c}{\begin{tabular}[c]{@{}c@{}}densely\\ annotated\end{tabular}} \\ \hline
    
    J-HMDB-21   &  one         &  one         & \xmark    & \cmark   \\ 
        UCF101-24   &  one or two  &  one         & \cmark    & \xmark  \\
    AVA   &  multiple  & multiple         & \xmark    & \cmark  \\
        \hline
    \end{tabular}}
    \caption{\textbf{Characteristics of the evaluated datasets}. Background people refers to no action being performed by people in some of the frames and which are not annotated}
    \label{tab:datasets_comparison}
    \vspace{-5mm}
\end{table}
\subsection{Datasets}
We evaluate our approach using two well-known spatial temporal action localization datasets namely UCF101-24 \cite{soomro2012ucf101} , J-HMDB-21 \cite{jhuang2013towards} and AVA\cite{Gu2018AVAAV}. The characteristics of each dataset is reported in Table \ref{tab:datasets_comparison}\\

UCF101-24 is a subset of UCF101\cite{soomro2012ucf101} action recognition dataset. UCF101-24 contains 3207 videos across 24 action classes with its corresponding annotations. The dataset is not densely annotated, with frames in a video where no action is performed. We conduct all our experiments on the first split of the dataset.

J-HMDB-21 is a subset of the HMDB-51 dataset \cite{kuehne2011hmdb}. It is a temporally trimmed dataset containing 928 videos with 21 action categories in daily life. We report our results on the first split.

AVA is a multi-action video detection dataset containing 80 atomic visual actions. The videos are densely annotated for 430 15-minute video clips resulting in 1.58 million action labels. Following Activity-Net Challenge, we reported the results on 60 action classes of the AVA dataset.

\subsection{Implementation Details}
We use the YOWO ~\cite{Kpkl2019YOWO} as the base architecture for all our experiments on spatial-temporal action localization. Following \cite{Kpkl2019YOWO} we used 3D-ResNext-101 as the 3D-CNN components of the YOWO architecture, and Darknet-19 \cite{redmon2017yolo9000} is used as the base model for the 2D-CNN backbone of the YOWO architecture. We used cosine classifier with two linear layers for the  frame-level classification head .We used two linear layers with sigmoid activation for the AVA dataset. We have used a batch size of $128$ video clips during training across all our experiments. We use the Adam gradient descent algorithm for our experiments with weight decay being set to $5e-4$. The learning rate is initialized with $0.001$ and reduced with a factor of $0.5$ after $3, 4, 5$ and $6$ epochs. We have used $\alpha=0.5, \beta=1, \gamma=2, \eta=0.5 $ and $\delta=1$ across our experiments on both J-HMDB-21 and UCF101-24 datasets. We have used $\gamma=0.5$ for AVA dataset while other hyper-parameters remains same. The end-to-end training is performed using PyTorch on AMD GPUs. Densely sampled 16 frames video clip is used to train the network. Following the usual practice ~\cite{Sohn2020Fixmatch} in SSL, we randomly
choose a certain proportion of labeled video samples across different action categories as a small labeled training set and discard the annotatations for the remaining data to create a large unlabeled training set.

Our framework is trained with different proportion of the labeled training data. We used $5\%$ and $10\%$ labeled data scenario for UCF101-24 and $20\%$ and $30\%$ labeled data scenario for J-HMDB-21 and AVA . We use horizontal flipping, random scaling, and random spatial cropping as the strong augmentation for our experiments. We used resized input images as the weakly augmented images in our experiments. We drop the classification head during the inference and use only the bounding-box head to make predictions on the test dataset. Since YOWO makes frame-level predictions, we used the linking algorithm defined in ~\cite{Peng2016MultiregionTR,Gkioxari2015FindingAT} to construct tublets for the video-level action detections. We evaluate our experiments using the well-known Video-mAP metric at the threshold of $0.1, 0.2$, and $0.5$ across all our experiments. We used Frame-mAP with $0.5$ IOU threshold as the annotations are sparsely provided for the AVA dataset with 1Hz.
\subsection{Baseline}
Following the usage of Mean-Teacher~\cite{Tarvainen2017Mean}  and Fix-Match ~\cite{Sohn2020Fixmatch}  in the semi-supervised object detection settings ~\cite{Tang2021HumbleTT,Liu2021UnbiasedTF,Zhou2021InstantTeachingAE}, we extend these baselines for the video domain with
YOWO as the backbone model. In Mean-teacher, we have two copies of the model, namely teacher and student, where
the teacher model is the EMA (exponential moving average) of the student model. Bounding boxes predicted by the teacher model which are above
the threshold are used as pseudo-bounding boxes for the student model. 

In Fix-Match, we also have two copies of the
model where the weights between the networks are shared. Weakly and strongly augmented versions of the samples are fed to the model.
Bounding boxes generated on the weakly augmented versions that are above the threshold are used as pseudo-bounding boxes for the strongly augmented version. In contrast to the baselines mentioned above and other work in
semi-supervised object detection ~\cite{Tang2021HumbleTT,Liu2021UnbiasedTF,Zhou2021InstantTeachingAE}, our approach extends a global video-action classifier in addition to a bounding box head. The global action classifier uses global context information to predict the video action, which is then used to select better pseudo-bounding boxes.

\subsection{Experiments and Comparison}
Tables \ref{tab:ucf101}-\ref{tab:jhmdb} show the performance of our method in comparison to the baselines and supervised approaches on different percentages of labeled data. We measure the performance of our work in pretrain and learning from scratch settings and report the Video-mAP performance.
\subsubsection{UCF101}
Table ~\ref{tab:ucf101} contains the results on $5\%$ and $10\%$ labeled data sampled from UCF101-24 dataset. In pretrained setting, the 3D and 2D backbone of YOWO is initialized with Kinetics and Darknet-19 weights. In table ~\ref{tab:ucf101}, we observe that our framework outperforms all the baseline approaches in both settings. We can see the huge performance improvement from $12.9$ to $26.0$ and $5.4$ to $9.4$ at the video-mAP threshold of $0.2$ and $0.5$ respectively for $10\%$ labeled data without any pretraining. We also see similar improvements in the pretrained setting. For the $5\%$ labeled data scenario,  the performance improved from $7.0$ to $13.8$ and $2.1$ to $4.6$ at the video-mAP threshold of $0.2$ and $0.5$, respectively. Our model achieves video-mAP of $40.5$ with pretrained weights while just $10\%$ of the labeled data. This clearly shows the superiority of our approach against the well-known SSL-based baselines.
\begin{table}[ht!]
\centering
\resizebox{\columnwidth}{!}{%
\begin{tabular}{l|l||cc|cc|cc}
\hline
& & \multicolumn{2}{c|}{V@0.1}& \multicolumn{2}{c|}{V@0.2}& \multicolumn{2}{c}{V@0.5} \\
\hline
Pretrain & Approach  & 5\% & 10\% &  5\% & 10\% & 5\% & 10\% \\
\hline
\multirow{4}{*}{\xmark} &
Supervised &15.4 &19.3  &7.0 &12.9 &2.1 &5.4  \\

& FixMatch &13.2 &22.3 &9.1 &14.0 &1.9 &4.5 \\
& Mean-Teacher &15.3 &24.9 &10.5 &16.9 &2.1 &5.1\\
& Ours &20.0 &34.9  &13.8 &26.0 &4.6 &9.4   \\
 
\hline
\hline
\multirow{4}{*}{\cmark} &
Supervised &61.1 &70.1  &52.2 &62.1 &27.0 &35.5  \\
& Mean-Teacher&66.3 &73.1 &60.1 &65.2 &34.1 &38.2 \\
& Fix-Match &64.3 &74.6  &54.6 &64.6 &32.3 &38.1\\
& Fully-Supervised ~\cite{Kpkl2019YOWO} &\multicolumn{2}{c|}{82.5}  &\multicolumn{2}{c|}{75.8} &\multicolumn{2}{c}{48.8} \\
& Ours &68.0 &74.0 &59.4 &66.1 &35.1  &40.5   \\

\hline
\end{tabular}
}
\vspace{1mm}
\caption{\small \textbf{Performance Comparison in UCF101-24 } Numbers show video-mAP at different threshold values for different percentages of the labeled data.We can clearly see our framework achieves best performance across all the baseline methods.
}
\label{tab:ucf101} 
\vspace{-4mm}
\end{table}
\subsubsection{J-HMDB-21}
Table \ref{tab:jhmdb} compares the video-mAP performance of the J-HMDB-21 dataset across different percentages ( $20\%$ and $30\%$) of labeled data. We can see the performance improvement achieved by our framework in both pretrain and learning from scratch settings. In the pretrained scenario, the 3D backbone is initialized with kinetics-101 weights finetuned on the HMDB-51 dataset, and the 2D backbone is initialized with Darknet-19 weights. 
\begin{table}[ht!]
\centering
\resizebox{\columnwidth}{!}{%
\begin{tabular}{l|l||cc|cc|cc}
\hline
& & \multicolumn{2}{c|}{V@0.1}& \multicolumn{2}{c|}{V@0.2}& \multicolumn{2}{c}{V@0.5} \\
\hline
Pretrain &Approach  & 20\% & 30\% &  20\% & 30\% & 20\% & 30\% \\
\hline
\multirow{4}{*}{\xmark} &
Supervised &7.2 &9.3  &6.2 &8.7 &1.1 &1.8   \\

&Fix-Match &9.6 &12.5 &9.0 &10.9 &2.4 &3.6\\
&Mean-Teacher &11.7 &12.7 &10.1 &11.8 &2.6 &4.1\\

& Ours &14.1 &10.9  &12.6 &9.8 &6.3 &6.6 \\
 \hline
\multirow{4}{*}{\cmark} &
Supervised &18.0 &26.0  &17.4 &25.3 &11.4 &18.9  \\
& Fix-Match &23.4 &27.8 &22.1 &27.4 &14.4 &20.8 \\
& Mean-Teacher &18.8 &33.0  &18.1 &32.3 &11.8 &24.6\\
& Fully-Supervised ~\cite{Kpkl2019YOWO} & \multicolumn{2}{c|}{-} &\multicolumn{2}{c|}{87.8}
&\multicolumn{2}{c}{85.7} \\
& Object-Prior\cite{Mettes2021ObjectPF} & \multicolumn{2}{c|}{32.1} &\multicolumn{2}{c|}{31.5} &\multicolumn{2}{c}{17.6}\\
& Ours &29.5 &38.3 &29.2 &38.0  &23.5  &34.8    \\

\hline
\end{tabular}
}
\vspace{1mm}
\caption{\small \textbf{Performance Comparison in J-HMDB-21 } Numbers show Video-mAP at different threshold values for different percentages of the labeled data.We can clearly see our framework achieves best performance across all the baseline methods.}
\label{tab:jhmdb} 
\vspace{-4mm}
\end{table}
Our approach performs better than the Mean-Teacher by $ \sim10$ points on $30\%$ labeled data. It also outperforms the Fix-Match by  $\sim9$ points on  $20\%$ labeled data scenario using pretrained weights. We also see the similar performance gains obtained by our framework without any pertaining. We achieve $6.6$ video-mAP$@0.5$ using $30\%$ labeled data compared to $1.8$ achieved by the supervised method. The performance gain achieved by our approach is $\sim 6$ times the supervised model on $20\%$ labeled data scenario. We can clearly observe that our method obtains superior performance against all the competing methods. 
\begin{table}[ht!]
\centering
\resizebox{\columnwidth}{!}
{%
\begin{tabular}{l|l||cc}
\hline
& & \multicolumn{2}{c}{F@0.5} \\
\hline
Pretrain &Approach  & 20\% & 30\% \\
\hline
\multirow{4}{*}{\cmark} &
Supervised &10.21 &11.69   \\

&Fix-Match  &10.32 &11.74\\
&Mean-Teacher  &10.13 &11.49\\
& Fully-Supervised ~\cite{Kpkl2019YOWO} &\multicolumn{2}{c}{17.9} \\
& Ours  &10.86 &12.38 \\
 \hline
\end{tabular}
}
\vspace{1mm}
\caption{\small \textbf{Performance Comparison in AVA } Numbers show Frame-mAP at $0.5$ threshold for different percentages of the labeled data. We can clearly see our framework achieves better performance than all the baseline methods.}
\label{tab:ava} 
\vspace{-4mm}
\end{table}
\subsubsection{AVA}
Table \ref{tab:ava} compares the frame-mAP performance of the AVA dataset across different percentages ( $20\%$ and $30\%$) of labeled data.
We can observe that our dual-guidance approach achieves better performance in comparison with baseline methods. We achieved $10.86$ and $12.38$ frame-mAP performance for the $20\%$ and $30\%$ labeled data scenarios, respectively, whereas the supervised method achieved only $10.21$ and $11.69$ for the $20\%$ and $30\%$ labeled data scenarios respectively. FixMatch only improves the performance marginally in both labeled data scenarios.
\subsection{Ablation}
We perform extensive ablation experiments to analyze our framework and its different components and hyperparameters. We perform these experiments on the UCF101-24 $10\%$ labeled data setting unless otherwise specified.

\subsubsection{Impact of Negative Sampling}
We analyze the impact of the absence of background frames in a video clip by performing an experiment in which unlabeled videos are trimmed. The background frames of the video clip have no associated action. We observe a significant drop in performance from $4.6$ to $2.2$ and $9.8$ to $6.1$ for $10\%$ and $5\%$ labeled data when using trimmed unlabeled data as indicated in table \ref{tab:background}. This clearly shows the importance of having background frames during network training.
\begin{table}[h!]
\centering
\resizebox{\columnwidth}{!}{%
\begin{tabular}{l||cc|cc|cc}
\hline
& \multicolumn{2}{c|}{V@0.1} &  \multicolumn{2}{c|}{V@0.2}& \multicolumn{2}{c}{V@0.5} \\
\hline
Approach  &  5\% & 10\%   &  5\% & 10\% & 5\% & 10\% \\
\hline
w/o Negative Sampling &15.9 &24.8  &10.9 &16.7  &2.2 &6.1\\
\hline
Ours &20.0 &35.2  &13.8 &26.0 &4.6 &9.4  \\

\hline
\end{tabular}
}
\vspace{1mm}
\caption{\small \textbf{Impact of Negative Samples on UCF101-24 dataset } Numbers show the Video-mAP across different threshold values. The drop in performance across different labeled data scenario indicates the importance of negative sampling during training.
}
\label{tab:background} \vspace{-2mm}
\end{table}

\begin{figure}
    \centering
    \includegraphics[width=\columnwidth]{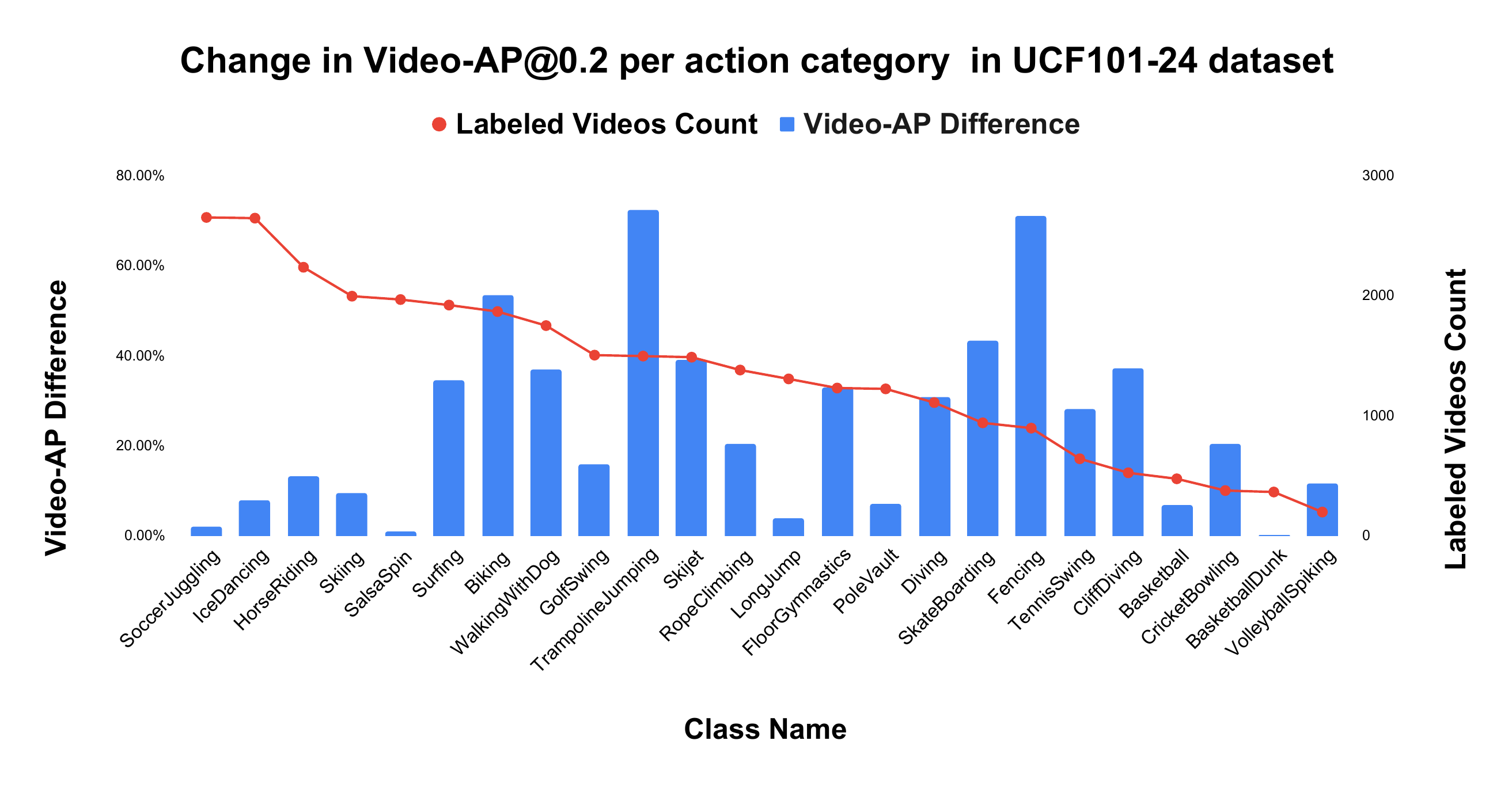}
    \caption{\textbf{Change in Video-mAP of our framework over supervised model on UCF101-24 $10\%$ labeled data}. Blue bars show the change in
video-mAP values on different action categories (sorted according to the number of training videos) of the UCF101-24 dataset under $10\%$ labeled scenario. Compared to Supervised model, our framework improves the Video-mAP performance of most classes in the UCF101-24 dataset. (Best viewed in color.)}
\label{fig:diff}\vspace{-2mm}
\end{figure}

\subsubsection{Analysis of different hyper-parameters}
\begin{figure}[t]
    \centering
    \includegraphics[width=\columnwidth]{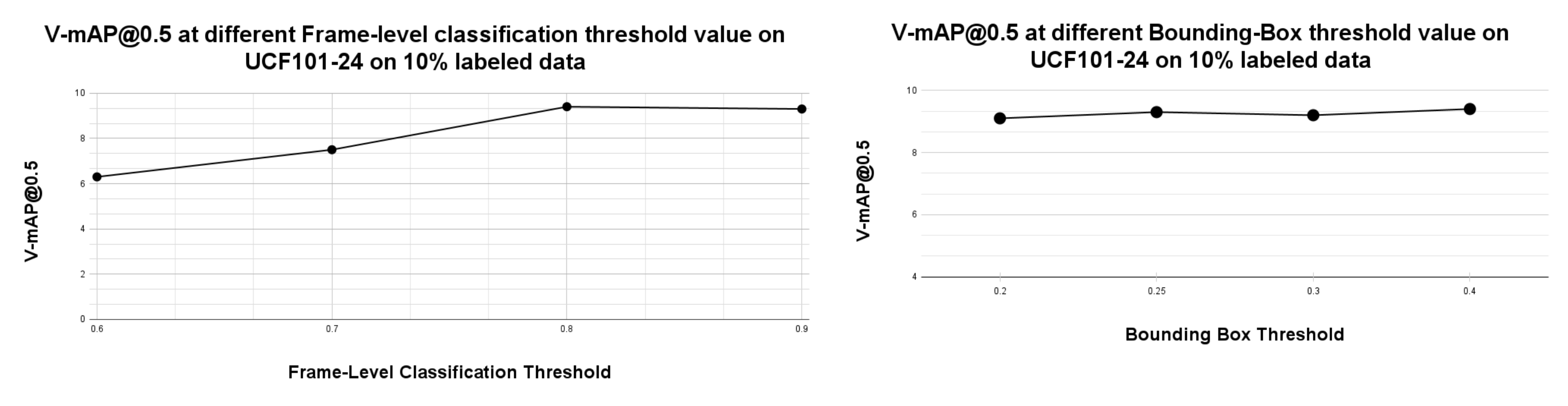}
    \caption{\textbf{Effect of different threshold on $10\%$ UCF101-24 labeled data setting without any pretrained weights.} \textbf{(a)} Effect of varying the frame-level classification threshold on test video-mAP values \textbf{(b)} Effect of varying Bounding Box threshold on test video-mAP values.}
    \label{fig:hyperparameters}
\end{figure}

We study the effect of frame-level classification and bounding-box thresholds on our framework. The impact of frame-level classification threshold can be seen in fig ~\ref{fig:hyperparameters}(a). Experimental results show the increase in video-mAP$@0.5$ values from $6.2$ to $9.4$ as the frame-level classification threshold increases from $0.6$ to $0.8$. The performance starts to saturate on further increase of the threshold. We used the bounding-box threshold of $0.4$ for our approach, which achieves maximum value as depicted in the figure ~\ref{fig:hyperparameters}(b).

\subsubsection{Classwise video-mAP}
We computed the video-mAP percentage difference between our framework and the supervised model trained on the UCF101-24 dataset using only $10\%$ labeled data. It helps to understand the effect of our approach on each action category present in the dataset. We observe a significant difference in video-map in most classes (in fig ~\ref{fig:diff}) as the number of labeled videos per class decreases, excluding the extremities where the number of labeled videos is either sufficiently high or low, reflecting a low video-mAP difference.

\subsubsection{Impact of Temporal Consistency}
We perform an experiment by removing the temporal component of our framework to analyze its effects.  The performance obtained without the temporal module is reported in the
table~\ref{tab:temporal_consistenccy_UCF}. We observe a drop in the frame-mAP$@0.5$ from $17.38$ to $15.76$ and $7.8$ to $7.0$ in both SSL settings of UCF101-24  dataset. We observe a similar drop in video-mAP on UCF101-24 from $9.4$ to $9.2$ and $4.6$ to $4.0$  in the absence of the temporal module  for $10\%$ and $5\%$ labeled data respectively. 
This shows the efficacy of the temporal module of our framework, which helps to gain improved performance by ensuring consistent representation across frames of the video clip.


\begin{table}[t!]
\centering
\begin{tabular}{l||cc||cc}
\hline
& \multicolumn{2}{c||}{V@0.5} & \multicolumn{2}{c}{F@0.5}\\
\hline
Approach  & 5\% & 10\% & 5\% & 10\% \\
\hline
w/o temporal &4.0 &9.2 &7.0 &15.76  \\
\hline
Ours &4.6 &9.4  &7.8 &17.38\\

\hline
\end{tabular}

\vspace{1mm}
\caption{\small \textbf{Impact of temporal consistency on UCF101-24 } The first columns show obtained Video-mAP$@0.5$
for different percentages of the labeled data. The last column shows the Frame-mAP$@0.5$ value for different SSL settings. The first row reports the numbers obtained without any temporal consistency during training. We can observe that having a temporal module in our framework helps to achieve better results.
}
\label{tab:temporal_consistenccy_UCF} \vspace{-2mm}
\end{table}

\begin{figure}
    \centering
    \includegraphics[width=\columnwidth]{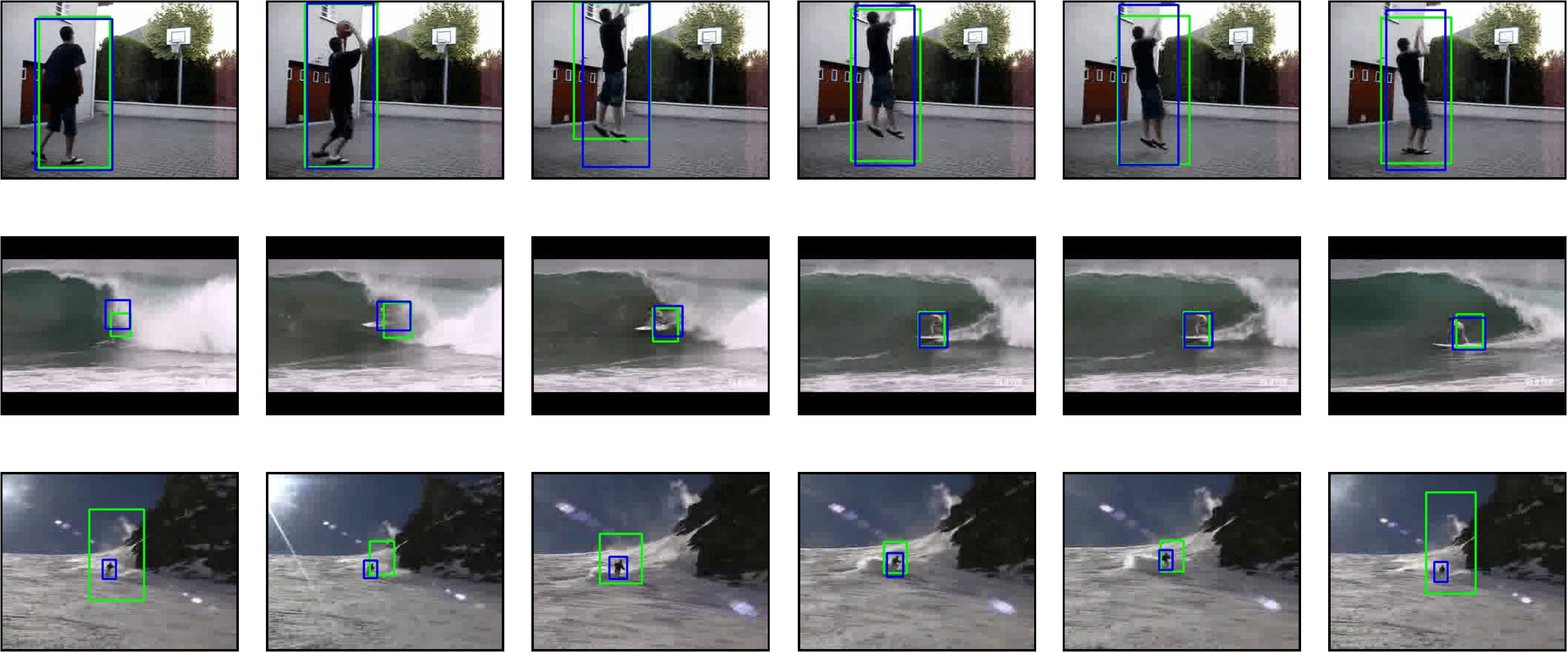}
    \caption{\textbf{Qualitative Examples of our approach on UCF101-24 test dataset} Blue colored box denotes the ground-truth while the green box denotes the bounding-box predicted by our framework at different frames of the video. Last row shows the failure case of our models. Our model is trained on UCF101-24 $10\%$ labeled dataset.}
    \label{fig:qualitative}
\end{figure}

\subsubsection{Qualitative Examples}
Figure \ref{fig:qualitative} shows qualitative examples of our framework on the UCF101-24 test dataset. As can be seen, our framework can correctly localize and recognize the action that is being performed in frames across the video. Our framework effectively uses the frame-level classification head and Bounding-Box head to select the good pseudo- bounding-boxes during training, reflecting better performance at that inference time.

\section{Conclusion}
We present a novel dual guidance framework for semi-supervised spatial-temporal action localization by using guidance from both frame-level classification head and bounding-box head to select the pseudo-labels for training the network. We employ frame-level classification head to select those pseudo-bounding boxes whose class-level prediction matches with its prediction. We demonstrate the effectiveness of our approach on the well-known spatial-temporal localization datasets, significantly outperforming the supervised methods.

\end{document}